\documentclass{article} 
\usepackage[preprint]{colm2026_conference}

\usepackage{microtype}
\usepackage{hyperref}
\usepackage{url}
\usepackage{booktabs}
\usepackage{multirow}
\usepackage{graphicx}
\usepackage{amsmath}
\usepackage[utf8]{inputenc}
\usepackage[T1]{fontenc}
\usepackage{comment}
\usepackage{cleveref}

\usepackage{lineno}

\definecolor{darkblue}{rgb}{0, 0, 0.5}
\hypersetup{colorlinks=true, citecolor=darkblue, linkcolor=darkblue, urlcolor=darkblue}

\title{Uniform Information Density-Based Preference Optimization}

\author{Yongyuan Li \& Yang Xu\thanks{Correspondence} \\
Department of Computer Science and Engineering\\
Southern University of Science and Technology\\
Shenzhen, Guangdong, China \\
\texttt{\{12432674,xuyang\}@sustech.edu.cn} \\
}

\begin{document}

\ifcolmsubmission
\linenumbers
\fi

\maketitle

\begin{abstract}
To enable large language models to generate outputs that align with human preferences, optimization algorithms based on the comparison between \emph{chosen} and \emph{rejected} output pairs have been developed and been proven effective. Some recent endeavors have utilized certain linguistically-derived assumptions about the preference pairs to guide the design of more precise optimization directions. 
However, human preferences are complex and hard to measure by simplistic criteria, and it is important to incorporate broader and more sophisticated insights about preferred vs. not-preferred language. 
In this study, we adopt a well-established psycholinguistic theory, Uniform Information Density, to develop a novel preference optimization algorithm, UIDPO. 
The basic assumption is that humans prefer to produce language in which information is uniformly distributed, because uniformity induces less cognitive processing effort. To implement this idea into the algorithm, we derive a compound metric for information uniformity by combining the global standard deviation and the local second-order difference, and integrate it to the objective of preference optimization. 
We compare UIDPO with SimPO, a previous state-of-the-art method, obtaining comparable and sometimes better performance on AlpacaEval2, ArenaHard, and MT-Bench. 
Additional linguistic analysis on complement clauses also shows that UIDPO can better reflect the tendencies of keeping information uniform in natural language than baselines.
Our results show that the subtle human preference over the information rate in language is a helpful source for achieving better human alignment for LLMs. The code repository is published here: \url{https://github.com/CLCS-SUSTech/UIDPO}.
\end{abstract}

\section{Introduction}

Preference optimization is an important step for developing high-performance and safe large language models (LLMs) that is aligned with human values \citep{leike2018scalableagentalignmentreward}. An aligned model can better interact with humans in a more ``helpful, honest, and harmless'' way \citep{askell2021generallanguageassistantlaboratory}. One common effective method for achieving this goal is reinforcement learning from human feedback (RLHF) \citep{ouyang2022traininglanguagemodelsfollow, christiano2023deepreinforcementlearninghuman, NEURIPS2020_1f89885d,zheng2023secretsrlhflargelanguage}, which applies  RL algorithms such as Proximal Policy Optimization \citep{schulman2017proximalpolicyoptimizationalgorithms} to finetune LLMs with joint objectives. However, RLHF relies on reward and value models, which are trained separately from the policy model, and thus induces high computational cost and brings optimization challenges \citep{casper2023openproblemsfundamentallimitations}.
To overcome the limits of RLHF and to simplify the optimization of PPO, different offline/off-policy algorithms are proposed. Direct Preference Optimization (DPO) \citep{rafailov2024directpreferenceoptimizationlanguage} is one of the successful alternatives. By re-parameterizing the general objective of PPO, DPO converts the original goal of optimizing the policy model to the optimization of an implicit reward function, which is a simpler, more stable, and effective alternative.

DPO is not without limitations, either. The implicit reward of DPO is the log ratio of the likelihood (probability) of a response between the current policy model and the \emph{reference} model. This reward is inconsistent with the inference-time metric for guiding the generation -- the log probabilities of tokens in the generated response according to the policy model. 
Simple Preference Optimization (SimPO) \citep{meng2024simposimplepreferenceoptimization} addresses this issue by replacing the implicit reward of DPO with a more straightforward form: the length-normalized log probability of response. This formulation more directly aligns the optimization objective with the inference-time metric. SimPO outperforms DPO over a broad range of instruction-following benchmarks, proving the plausibility of its reward formulation. 

However, although SimPO provides a simplified methodological framework to learn direct human preferences from response pairs, more fine-grained guidance is still needed in order to maintain a good generation performance from the outcome model. This is because SimPO is reference-free, which means it lacks constraints on the optimization direction -- it only understands some responses are preferred over others, but does not recognize which linguistic aspects it should focus on. This limitation is addressed by some recent feature-based controllable optimization methods, such as FPO \citep{yin2025constrainalignmentsparseautoencoders}. 
Our work follows this general direction by incorporating a theoretically motivated linguistic property into prefernece optimization. Specifically, we draw on a well-established theory about human preference over how information should be distributed temporally language -- the uniform information density (UID) hypothesis \citep{NIPS2006_c6a01432,FLORIANJAEGER201023}, according to which language users tend to distribute information relatively uniformly within a linguistic sequence (sentence, utterance, or longer discourse). 
We therefore formulate an implicit reward based on the distributional \emph{variance} of token-level information within a response, rather than the sheer average amount, as in SimPO.

As a brief overview of this study, we further motivate the design and formulate a novel preference optimization reward by quantifying the uniformity of information in \Cref{sec:UIDPO}. We demonstrate that the method reaches SoTA performance in common LLM alignment benchmarks and reveals interesting human-alignment results in relevant linguistic analysis in \Cref{sec:Experiments}. We believe our work is a meaningful attempt to bring a long-standing psycholinguistic observation to the realm of LLM post-training. 


\section{UIDPO: Uniform Information Density-based Preference Optimization}\label{sec:UIDPO}

\subsection{UID and Human Preference}
In the terminology of computational linguistics, the amount of information encoded in a linguistic unit can be measured with \emph{surprisal} (negative log probability) \citep{hale-2001-probabilistic}. It has clear psychological meanings, as the cognitive load of language processing (measured by reading times, eye-movement etc.) is correlated to the surprisal values \citep{Shain2024surprisalReadingTime}. The average amount of surprisal within a sequence is also termed as \emph{information density}. The theory of uniform information density (UID) \citep{NIPS2006_c6a01432, informationdensityanddependency} makes broader predictions about human preferences over linguistic structures (e.g., word orders, syntactic structures) based on the degrees of uniformity of information, and it assumes that uniformly distributed information have a good balance between the efficiency and the reliability in transmission and therefore is preferred by human. To briefly summarize, UID provides an information-theoretic explanation to why certain humans preferences exist in natural language.

The natural association between UID and human preference has drawn attention in recent LLM studies. Computational metrics are developed to measure the ``degree'' of UID in LLMs generations \citep{informationdensityanddependency,meister2021revisitinguniforminformationdensity}. It is found that high-quality LLM responses for reasoning tasks tend to present more UID features than low-quality ones \citep{gwak2026revisitinguniforminformationdensity}, which  motivates the idea of this study to use UID feature as an implicit reward for preference optimization. 

\subsection{Metrics for quantifying UID} 
\begin{figure}[h]
    \centering
    \includegraphics[width=.8\linewidth]{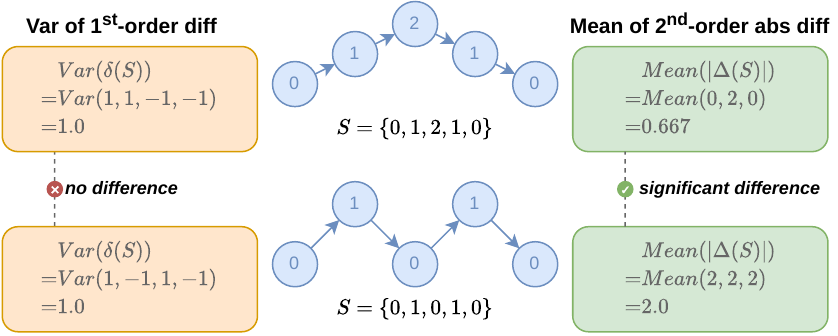}
    \caption{Comparison of metrics for local information uniformity.}
    \label{fig:metrics_smoothness}
\end{figure}

We make changes to previous formulation of UID to adjust for the preference optimization task. Before our study, the degree UID is defined as two components \citep{informationdensityanddependency,gwak2026revisitinguniforminformationdensity}: global uniformity and local uniformity, which are formulated as follows:

\begin{equation}
    U_{global}(y) = \mathrm{Var}(\log\pi_{\theta}(y_1 \mid y_0),..., \log\pi_{\theta}(y_T \mid y_{<T})),
\end{equation}
\begin{equation}
    U_{local}(y) = \mathrm{Var}(\delta_1,..., \delta_N),
\end{equation}

where $\delta_k = \log\pi_\theta(y_k \mid y_{<k}) - \log\pi_\theta(y_{k-1} \mid y_{<k-1})$, $\log\pi_\theta(y_k\mid y_{<k})$ is the log probability (log-softmax of model output logits) of ground truth token $y_k$ at position $k$ given all the previous ground truth $y_{<k}$, $\mathrm{Var}$ means variance, $T$ is the length of a trajectory, and $\theta$ is the trainable parameters of a language model.

With this formulation, the ``uniformity'' of information is quantified with the variance, and lower variance indicates better uniformity. The idea of separating global and local uniformity is to describe the information pattern of long texts more precisely. It is common that longer texts often switch between topics that have different information density levels. For instance, a professional speech with high information density might start with some greetings with low information density.

The original UID assumption does not consider contents with multiple topics. It expect information density to be uniform throughout the text, resulting in low uniformity variance both globally and locally. However, in most real-world scenarios, UID is more observable within discourse units of fixed lengths, and may exhibit jumping changes near discourse boundaries \citep{xu2018information,tsipidi-etal-2024-surprise}. Actually it is rare to observe the information density staying uniform throughout a long scope, but rather a constantly changing pattern that results in a ubiquitous periodicity \citep{tsipidi-etal-2025-harmonic,ou2025periodicity}. This fact is also noticed and discussed in recent UID-related work on reasoning trajectories of LLMs \citep{gwak2026revisitinguniforminformationdensity}. \textbf{Therefore, we revise the original assumption to encourage large global information density range, but restrict dramatic local information density change}.

Given the new assumption, we need to reformulate the global and local uniformity. We initially followed the above formulations for pilot experiments, but the results were not ideal: the reward gap between chosen responses and rejected responses is small; loss and reward stopped converging early; the output logits did not change significantly; and the accuracy of chosen responses being assigned higher reward scores is only slightly over a random level.
Based on the pilot results, we hypothesize that it is necessary to look for a fine-grained ``division of duties'' between the global and local uniformity. We propose the new formulation for local uniformity as follows:

\begin{equation}
    U_{local}(y) = \frac{1}{T-2}\sum^{T-2}_{t=1}\sqrt{\Delta_t^2 + \epsilon},
\end{equation}

where $\Delta_k = \log\pi_\theta(y_k \mid y_{<k}) + \log\pi_\theta(y_{k+2} \mid y_{<k+2}) - 2\log\pi_\theta(y_{k+1} \mid y_{<k+1})$, and $\epsilon$ is a small constant to avoid division by zero in gradient. Here, $\sqrt{\Delta_t^2 + \epsilon}$ is a common approximation for $\left|\Delta_t\right|$.

We refer to this definition as the mean of second-order absolute differences of log probabilities. It is also known as the second term of the $\ell_1$ trend filter \cite{l1trendfilter}. It has two advantages over the previous definition, i.e., variance of first-order differences: First, our assumption prefers information density to be locally linear and have sparse curvature, instead of being globally smooth, since we allow large information density range globally but restrict local information density to be uniform. This property is also called ``Piecewise Linearity''. The second term of the $\ell_1$ trend filter is proposed to exactly fit our need. Second, mean value is much less sensitive to outliers than variance, making it a more accurate reward function. An example of the first advantage is shown in Figure \ref{fig:metrics_smoothness}. Variance of the first-order differences has no preference between the two sequences, while mean of the second-order absolute differences significantly distinguish between them. According to our assumption, we prefer the first sequence. Therefore, the mean of second-order absolute differences better fits our need.

Now that the local uniformity is in first power, we change the global uniformity to its square root, to match the power. That is, to replace the variance with standard deviation:

\begin{equation}
    U_{global}(y) = \sqrt{\mathrm{Var}(\log\pi_{\theta}(y_1 \mid y_0),..., \log\pi_{\theta}(y_T \mid y_{<T}))}.
\end{equation}

\subsection{UIDPO Objective}

With the new assumption and the new uniformity formulations, we define the UIDPO reward as:

\begin{equation}
    r_{UIDPO}(x, y) = \beta(\alpha U_{global}(y\mid x) - (1 - \alpha)U_{local}(y\mid x)),
\end{equation}

where $x$ and $y$ are prompt and response, $\beta$ is a constant that controls the scaling of the reward difference, and $\alpha$ is a constant that controls the blending ratio of global and local uniformity metrics. Then we construct the learning objective of UIDPO on the basis of SimPO, by replacing the SimPO reward with UIDPO reward. The final loss is written as:

\begin{equation}
    \begin{split}
        \mathcal{L}(\pi_\theta) =& - \mathbb{E}_{(x, y_w, y_l)\sim \mathcal{D}}\big[\log\sigma\left(r_{UIDPO}(x, y_w) - r_{UIDPO}(x, y_l) - \gamma\right)\big], \\
        =& - \mathbb{E}_{(x, y_w, y_l)\sim \mathcal{D}}\big[\log\sigma(\beta\alpha(U_{global}(y_w\mid x) - U_{global}(y_l\mid x)) \\
        & -\beta(1 - \alpha)(U_{local}(y_w\mid x) - U_{local}(y_l\mid x)) - \gamma\big],
    \end{split}
\end{equation}

where $(x, y_w, y_l)$ are preference pairs consisting of the prompt, the chosen response, and the rejected response from the preference dataset $\mathcal{D}$, and $\gamma$ is a constant proposed by SimPO that ensures the reward gap between the two responses.

\section{Experiment Setup}

To be comparable with the SimPO experiment results, we follow the recommended setups specified in the SimPO code repositories. We also include ablation experiments on key hyper-parameters of UIDPO, and several interpretability experiments.

\paragraph{Models.} All of our experiments are conducted over two instruct models from two families: Llama3-8B-Instruct\footnote{\href{https://huggingface.co/meta-llama/Meta-Llama-3-8B-Instruct}{meta-llama/Meta-Llama-3-8B-Instruct}} \citep{llama3modelcard} and Mistral-7B-Instruct\footnote{\href{https://huggingface.co/mistralai/Mistral-7B-Instruct-v0.2}{mistralai/Mistral-7B-Instruct-v0.2}} \citep{jiang2023mistral7b}.
We exclude the experiments on base models because according to SimPO experiment results, instruct models always get better performance under the same setup. Therefore, including only the instruct setup is enough to compare UIDPO with SimPO and indirectly with all the other preference optimization methods included in the SimPO experiments.

\paragraph{Dataset.} We use \href{https://huggingface.co/datasets/princeton-nlp/llama3-ultrafeedback-armorm}{princeton-nlp/llama3-ultrafeedback-armorm} as the preference data. It is not the same dataset reported in the SimPO paper, but the one with the best performance in the SimPO code repository. We choose this dataset to compare UIDPO with the best SimPO scores. It contains around 60k training samples and 2k testing samples, following the original splits\footnote{\href{https://huggingface.co/datasets/HuggingFaceH4/ultrafeedback_binarized}{HuggingFaceH4/ultrafeedback\_binarized}}\citep{cui2023ultrafeedback}. The responses in this dataset are generated with Llama-3-8B-Instruct and annotated with the reward model ArmoRM-Llama3-8B-v0.1\footnote{\href{https://huggingface.co/RLHFlow/ArmoRM-Llama3-8B-v0.1}{RLHFlow/ArmoRM-Llama3-8B-v0.1}}\citep{wang2024interpretable}. However, this leads to an inconsistency issue when the dataset is used to train Mistral-7B-Instruct. Since we are mainly comparing Llama-3-8B-Instruct, and this issue affects both UIDPO and SimPO, we decide to tolerate it and include results of Mistral-7B-Instruct for reference.

\begin{table}[h]
    \centering
    \begin{tabular}{lcccc}
        \toprule
        Benchmark & Baseline Model & Judge Model & Metric \\
        \midrule
        AlpacaEval 2 & GPT-4 1106 Preview & GPT-4 1106 Preview & LC \& raw win rate \\
        AlpacaEval 2* & GPT-4 1106 Preview & Llama-3.3 70B IT & LC \& raw win rate \\
        Arena-Hard v0.1 & GPT-4 0314 & Llama-3.3 70B IT & Win rate \\
        MT-Bench & -- & Llama-3.3 70B IT & 0-10 rating \\
        \bottomrule
    \end{tabular}
    \caption{Detailed benchmark setups. Llama-3.3 70B IT refers to Llama-3.3 70B Instruct Turbo. LC means length controlled win rate. For AlpacaEval2 and Arena-Hard, the baseline responses to be compared with the subject are generated with the baseline model. All the comparison and rating are conducted by the judge model. Note the difference between AlpacaEval 2 and AlpacaEval 2*.}
    \label{tab:benchmarks}
\end{table}

\paragraph{Benchmarks.} We evaluate our models with three popular open-ended instruction-following benchmarks: AlpacaEval 2 \citep{alpaca_eval, dubois2024length}, Arena-Hard v0.1 \citep{arenahard2024, li2024crowdsourced}, and MT-Bench \citep{zheng2023judging}. AlpacaEval 2 is known to have a spearman correlation of 0.98 with \href{https://huggingface.co/spaces/lmarena-ai/arena-leaderboard}{Chatbot Arena} \citep{chiang2024chatbotarenaopenplatform} and to be less biased on longer responses. It consists of 805 questions from 5 datasets and judges with large language models (LLM-as-judge). Arena-Hard is another LLM-as-judge benchmark that consists of curated prompts sourced from Chatbot Arena. Both of the above two benchmarks compare the model's responses with baselines'. MT-Bench includes 8 categories with 80 questions and evaluates models by directly rating the responses with a judge model.
The detailed setup is listed in Table \ref{tab:benchmarks}. Due to limited time and token quota, only part of the experiments are judged by GPT-4. In the following result tables, scores reported under ``AlpacaEval 2'' are judged by GPT-4, which is the same with SimPO experiments and can be directly compared. Scores reported under ``AlpacaEval 2*'', ``Arena-Hard v0.1'', and ``MT-Bench'' are judged by Llama-3 70B. For these benchmarks, we reproduce the best performing SimPO with exactly the same parameters reported, and run under the same setup to compare with UIDPO. We also run the SFT models under all the benchmark setups that are different from SimPO, to complete the baselines.

\section{Experiment Results}\label{sec:Experiments}

\subsection{Main Results and Ablations}

\begin{table}[h]
\small
    \centering
    \begin{tabular}{lccccccc}
        \toprule
        \multirow{2}{*}{\textbf{Method}} & \multicolumn{2}{c}{\textbf{AlpacaEval 2}} & \multicolumn{2}{c}{\textbf{AlpacaEval 2*}} & \textbf{Arena-Hard v0.1} & \textbf{MT-Bench} \\
        \cmidrule(lr){2-3}\cmidrule(lr){4-5}
        & LC(\%) & WR(\%) & LC(\%) & WR(\%) & WR(\%) & Score(0-10) \\
        \midrule
        Instruct & 26.0 & 25.3 & 45.80 & 43.66 & 40.8 & 7.2 \\
        SimPO v0.2 & 53.7 & 47.5 & -- & -- & -- & -- \\
        \midrule
        SimPO reprod. & \underline{\textbf{51.9}} & \underline{\textbf{47.4}} & 60.8 & 56.1 & \underline{\textbf{66.6}} & 7.3 \\
        UIDPO & 49.0 & 44.8 & \underline{\textbf{63.2}} & \underline{\textbf{56.2}} & 62.7 & \underline{\textbf{7.5}} \\
        \bottomrule
    \end{tabular}
    \caption{Benchmark scores of UIDPO and SimPO. SimPO v0.2 is the best performing SimPO model for reference. Its scores are only available under the ``AlpacaEval 2'' setup. The row of Instruct is from the Llama-3-8B-Instruct baseline without any post-training. The row of SimPO reprod. is from the reproduced model under exactly the same parameters of SimPO v0.2. LC is the length-controlled win rate, and WR is the raw win rate. Better scores are in bold and underlined.}
    \label{tab:mainresult}
\end{table}

\paragraph{UIDPO is comparable to SimPO.} UIDPO achieves comparable performance to SimPO under all benchmarks and settings. As shown in Table \ref{tab:mainresult}, UIDPO and our reproduced SimPO each wins half of the benchmarks with minor differences in scores. In the ablation experiment, the observed error in training and AlpacaEval 2 is around $\pm 2.5$, so the differences in AlpacaEval 2 is likely within the random error. More AlpacaEval 2 results from SimPO paper can be found in Appendix \ref{appendix:moreresults}. The responses of UIDPO has slightly better length statistics, which are discussed in Appendix \ref{appendix:statistics}.

\begin{table}[h]
\small
    \centering
    \begin{tabular}{ccccccccc}
        \toprule
        & \multicolumn{4}{c}{Llama-3-8B-Instruct} & \multicolumn{4}{c}{Mistral-7B-Instruct} \\
        \cmidrule(lr){2-5}\cmidrule(lr){6-9}
        \textbf{$\alpha$} & \textbf{LC(\%)} & \textbf{WR(\%)} & \textbf{SE} & \textbf{avg. length} & \textbf{LC(\%)} & \textbf{WR(\%)} & \textbf{SE} & \textbf{avg. length} \\
        \midrule
        0.25 & \underline{\textbf{63.2}} & \underline{\textbf{56.2}} & 1.75 & 1799 & \underline{\textbf{34.3}} & 23.2 & 1.49 & 1550 \\
        0.25 orig. & 21.0 & 25.0 & 11.4 & 1893 & 31.29 & \underline{\textbf{33.29}} & 1.66 & 2895 \\
        \midrule
        1.00 & 32.0 & 27.1 &  1.57 & 1743 & \underline{\textbf{9.91}} & 4.66 & 0.74 & 812 \\
        1.00 orig. & \underline{\textbf{40.8}} & \underline{\textbf{35.8}} & 1.69 & 1829 & 5.4 & \underline{\textbf{7.6}} & 0.93 & 7221 \\
        \midrule
        Inst. & 45.8 & 43.7 & 1.75 & 1946 & 19.24 & 13.42 & 1.20 & 1500 \\
        \bottomrule
    \end{tabular}
    \caption{AlpacaEval 2* scores of UIDPO models following the original UID assumption and the UIDPO assumption. Rows with ``orig.'' are from models trained under the original UID assumption. The Inst. row is from the Instruct models without any post-training. 
    SE is the standard error (normalized by $N-1$) of the win rate. 
    Avg. length is the average token length of responses. Better scores in the same group are in bold and underlined.}
    \label{tab:ablationassumption}
\end{table}

\paragraph{UIDPO assumption works, and original assumption doesn't.} Before presenting complete ablations on $\alpha$, we first compare rewards formulated following original and UIDPO assumptions, with two special $\alpha$, $0.250$ and $1.000$. The former is the best performing parameter, and the latter makes the UIDPO reward completely depend on global uniformity. Since the new UID assumption of UIDPO changes the optimization direction of the global uniformity, the comparison between both directions is needed. The AlpacaEval 2* scores of models trained under the two assumptions and with two different $\alpha$s are listed in Table \ref{tab:ablationassumption}.

When $\alpha=1.00$, both original UID and UIDPO assumptions fail to improve the instruct model of both LLama-3-8B-Instruct and Mistral-7B-Instruct. All scores with $\alpha=1.00$ are lower than that of the Instruct model. On Llama-3-8B-Instruct, the original assumption causes less degradation in terms of LC win rate. On Mistral-7B-Instruct, the performance reverses. This shows that UIDPO does not work with only global uniformity, irrelevant of the optimization direction. We can infer that the standard deviation of log probabilities is too rough to solely reward the policy model. An evidence is the average length on Mistral-7B-Instruct. The huge gap between $812$ and $7221$ indicates that UIDPO with only global uniformity is probably hacked by length. Lengthy nonsense can effectively decrease the standard deviation of log probabilities, and short content does the reverse. These results prove the necessity to separate global and local uniformity.

When $\alpha=0.25$, UIDPO always improves the performance, while UIDPO with original UID assumption worsen the performance on Llama-3-8B-Instruct. Its standard error of the win rate is significantly larger under the original UID assumption. This shows that the reward with original UID assumption is not a valid regularization on Llama-3-8B-Instruct, making the performance of different responses varies severely. On Mistral-7B-Instruct, UIDPO with original UID assumption improves the performance. It hacks the raw win rate again with sharply increased average length. But its gap of length-controlled win rate with UIDPO assumption is small, and the standard error is also reasonable. This shows that in some cases, the original UID assumption could work. However, the necessary condition for it to work is currently unclear. Hence we consider the new UIDPO assumption is more stable than the original UID assumption, on preference optimization tasks. With above observations, we conclude that the UIDPO assumption successfully adapts the original UID assumption to preference optimization tasks.

\begin{table}[h]
\small
    \centering
    \begin{tabular}{ccccccccc}
        \toprule
        & \multicolumn{4}{c}{Llama-3-8B-Instruct} & \multicolumn{4}{c}{Mistral-7B-Instruct} \\
        \cmidrule(lr){2-5}\cmidrule(lr){6-9}
        \textbf{$\alpha$ / $\frac{\gamma}{\beta}$} & \textbf{LC(\%)} & \textbf{WR(\%)} & \textbf{SE} & \textbf{avg. length} & \textbf{LC(\%)} & \textbf{WR(\%)} & \textbf{SE} & \textbf{avg. length} \\
        \midrule
        0.000 & 59.9 & 52.7 & 1.92 & 1839 & 34.0 & \underline{\textbf{32.4}} & 1.65 & 1970 \\
        0.125 & 60.7 & 54.2 & 1.76 & 1847 & 34.1 & 30.7 & 1.63 & 1836 \\
        0.250 & \underline{\textbf{63.2}} & \underline{\textbf{56.2}} & 1.75 & 1799 & \underline{\textbf{34.3}} & 23.2 & 1.49 & 1550 \\
        0.375 & 62.3 & 55.3 & 1.75 & 1816 & 28.0 & 21.4 & 1.45 & 1616 \\
        0.500 & 41.0 & 33.6 & 1.66 & 1745 & 25.7 & 20.0 & 1.41 & 1621 \\
        0.625 & 36.6 & 30.3 & 1.62 & 1736 & 28.1 & 16.5 & 1.31 & 1352 \\
        \midrule
        0.1 & 58.6 & 50.8 & 1.76 & 1781 & \underline{\textbf{34.3}} & 23.2 & 1.49 & 1550  \\
        0.3 & \underline{\textbf{63.2}} & \underline{\textbf{56.2}} & 1.75 & 1799 & 32.6 & \underline{\textbf{24.0}} & 1.51 & 1566 \\
        0.5 & 62.1 & 55.2 & 1.75 & 1828 & 30.6 & 21.1 & 1.44 & 1519 \\
        0.7 & 60.4 & 54.3 & 1.75 & 1840 & 21.9 & 23.0 & 1.48 & 4229 \\
        0.9 & 62.4 & 55.9 & 1.75 & 1825 & 25.8 & 18.1 & 1.36 & 1494 \\
        1.1 & 59.4 & 53.7 & 1.76 & 1856 & 30.9 & 21.2 & 1.44 & 1529 \\
        1.3 & 59.6 & 52.2 & 1.76 & 1793 & 26.4 & 17.8 & 1.35 & 1477 \\
        \midrule
        Inst. & 45.8 & 43.7 & 1.75 & 1946 & 19.24 & 13.42 & 1.20 & 1500 \\
        \bottomrule
    \end{tabular}
    \caption{AlpacaEval 2* scores of models trained with different $\alpha$ and $\gamma$. The top six rows are ablations on $\alpha$, and the rests are ablations on $\gamma$ (controlled with $\frac{\gamma}{\beta}$). 
    Best scores in the same group are in bold and underlined.}
    \label{tab:ablationparams}
\end{table}

\paragraph{Ablations on $\alpha$ and $\gamma$.} We look into the influence of the key parameter $\alpha$ and $\gamma$. $\beta$ is also crucial. The role of $\beta$ and its influence on the training has been fully discussed in the SimPO paper. Our early experiments with small stride ablation over $\beta$ is based on a feasible initial value of $\beta$ given by SimPO, and the performance is consistent. Larger stride results in sudden changes in convergence, because the influence is probably not linear. Hence it is not included in the ablation study due to limited time. 
$\alpha$ is tuned directly, while $\gamma$ is controlled by tunning $\frac{\gamma}{\beta}$. The default parameters for Llama-3-8B-Instruct are $\alpha = 0.25, \beta = 10, \frac{\gamma}{\beta} = 0.3$. The default parameters for Mistral-7B-Instruct are $\alpha = 0.25, \beta = 2.5, \frac{\gamma}{\beta} = 0.1$. Table \ref{tab:ablationparams} shows the alpacaEval 2* scores under different $\alpha$ and $\gamma$.

When $\alpha = 0.000$, the performances of UIDPO models are better than that of the Instruct models, but worse than that of the UIDPO models under $\alpha = 0.250$. This indicates that the new local uniformity definition of UIDPO is effective, even without the global uniformity. Also, the looser restriction of local uniformity could relax the model from hacking with length, so that Mistral-7B-Instruct does not show abnormally large average length like it does with the original global uniformity.
The best length-controlled win rate is obtained with $\alpha = 0.250$ for both Llama-3-8B-Instruct and Mistral-7B-Instruct, showing consistency and stability in parameters across model backbones. Since $\alpha$ controls the proportion of mixture of the global and the local uniformity, and both uniformity are numerically in the same power, we believe that $\alpha = 0.250$ is generally a good parameter for all setups.
On Mistral-7B-Instruct, the average length of responses decreases as $\alpha$ increases. The raw win rate decreases with the average length. This again shows that Mistral-7B-Instruct has larger tendency of hacking with length. As a contrast, the average length of Llama-3-8B-Instruct is very stable across all $\alpha$ setups. This difference might due to the inconsistency between Mistral-7B-Instruct and the preference data generated with Llama-3-8B-Instruct.
$\gamma$ is discussed in SimPO paper, and our ablations exhibit the same observation. Increasing the reward gap between chosen and rejected responses causes the flattening of the distribution, which can be confirmed with a decreasing log probability (figure in Appendix \ref{appendix:logpgamma}). This initially improves performance but can eventually lead to model degeneration.

\subsection{Linguistic Analysis}

\begin{figure}
    \centering
    \includegraphics[width=\linewidth]{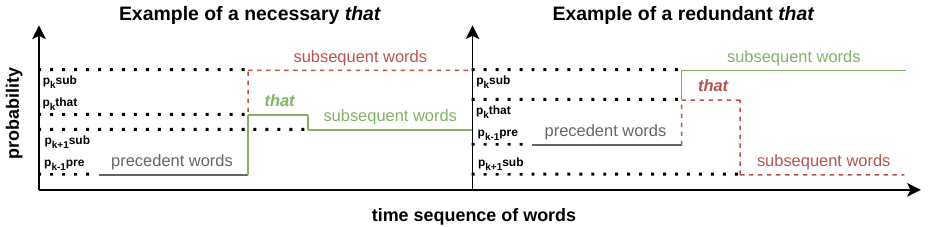}
    \caption{Exhibition of our linguistic analysis method. Assuming the conjunction word ``\textit{that}'' is at position $k$, whose probability is $p_k(\textit{that})$. The probabilities of words before and after are $p_{k-1}(\textit{pre})$ and $p_{k+1}(\textit{sub})$, respectively. The probability of a subsequent word at position $k$ immediately following the preceding (thus, omitting \emph{that}) is $p_{k}(\textit{sub})$. Left curves showcase a necessary \textit{that}. Right curves showcases a redundant \textit{that}.}
    \label{fig:linguisticanalysis}
\end{figure}

To verify the linguistic advantage of UIDPO, we examine how well the finetuned models can quantify the optional \textit{that}-mentioning in English complement clauses \citep{FLORIANJAEGER201023}, which is a well-established evidence of UID in language use. As shown in Figure \ref{fig:linguisticanalysis}, for a given model, we compute the negative log probabilities of the words surrounding the conjunction \textit{that}. We define the value of using \textit{that} as:

\begin{equation}
    \text{Value}(that)=\left|p_{k-1}(\textit{pre}) - p_k(\textit{sub})\right| - \\ \Big[\left|p_{k-1}(\textit{pre}) - p_k(\textit{that})\right| + \left|p_k(\textit{that}) - p_{k+1}(\textit{sub})\right|\Big].
\end{equation}

Intuitively, this quantity measures the difference between \emph{with} and \emph{without} using ``\emph{that}'' in the total amount of local information variance. To the left of subtraction is the absolute information change between the preceding and subsequent words, without using ``\emph{that}''; to the right is the sum of information changes with ``\emph{that}'' inserted in-between. 
For a given sentence, if $\text{Value}(that)>0$, then it indicates inserting \textit{that} as a complementizer makes the information density transition smoother, and thus it is labeled as a necessary \textit{that}. Otherwise, it is labeled as a redundant \textit{that}. 
Next, we compute the probabilities at position $k$, $p_k(\textit{that})$ and $p_k(\textit{sub})$. For a necessary \emph{that}, we expect to see $p_k(\textit{that})>p_k(\textit{sub})$, and if so, we mark it as ``present'' (correct), meaning that it is likely to be sampled in generation; otherwise, we mark it as ``absent'' (mistaken). For a redundant \emph{that}, the mapping is switched: $p_k(\textit{that})<p_k(\textit{sub})$ maps to ``absent'' (correct); otherwise maps to ``present'' (mistaken).
The counts of these four cases are presented in \Cref{tab:conjunctionthat}.

\begin{table}[h]
    \centering
    \begin{tabular}{lccc}
        \toprule
        \textbf{Metric} & \textbf{SFT} & \textbf{SimPO} & \textbf{UIDPO} \\
        \midrule
        \# of necessary \textit{that}: \\
        present/correct ($\uparrow$) & 495 (0\%) & 617 (24.65\%) & \textbf{676 (36.57\%)} \\
        absent/mistaken ($\downarrow$) & 186 (0\%) & \textbf{181 (-2.69\%)} & 191 (2.69\%) \\
        \midrule
        \# of redundant \textit{that}: \\
        absent/correct ($\uparrow$) & \textbf{348 (0\%)} & 343 (-1.44\%) & 345 (-0.86\%) \\
        present/mistaken ($\downarrow$) & 3719 (0\%) & 3607 (-3.01\%) & \textbf{3536 (-4.92\%)} \\
        \midrule
        Total \# of \textit{that} \\
        correct ($\uparrow$) & 843 (0\%) & 960 (13.88\%) & \textbf{1021 (21.12\%)} \\
        mistaken ($\downarrow$) & 3905 (0\%) & 3788 (-3.00\%) & \textbf{3727 (-4.56\%)} \\
        \bottomrule
    \end{tabular}
    \caption{The numbers of necessary and redundant \textit{that} (present vs. absent). All three models have Llama-3-8B-Instruct backbone. Up arrow means the larger the better, and down arrow means the opposite. The percentages in the brackets are the rates of change relative to the SFT baseline. Best scores are in bold.}
    \label{tab:conjunctionthat}
\end{table}

We collect the sentences containing \textit{that} complement clauses from the Switchboard corpus \citep{godfrey1992switchboard}, following the same \texttt{tgrep2} \citep{rohde2005tgrep2} pattern-based extraction procedure used in \citet{FLORIANJAEGER201023}'s analysis. 
4817 sentences are extracted in total, from which we count the numbers of necessary vs. redundant \textit{that} being present or absent, using probabilities computed from the SFT, SimPO, and UIDPO (Llama-3-8B-Instruct backbone) models, respectively. From Table \ref{tab:conjunctionthat}  we can see that UIDPO greatly improves the number of correct \textit{that}, and also reduces the number of mistaken \textit{that} by a smaller margin -- \textbf{UIDPO outperforms SFT and SimPO on both metrics}.

Using the sentence $S_1:\,\textit{My boss confirmed that we were absolutely crazy}$ as an example, and let $c$ denote the context \textit{My boss confirmed}, we examine the probabilities of \emph{that} and the alternative continuation \emph{we}. 
When a UIDPO-finetuned model (Llama-3-8B-Instruct, $\alpha=0.250$) is used to estimate probabilities, we have $p^{\text{UIDPO}}(\textit{that}\mid c)=0.4996$ and $p^{\text{UIDPO}}(\textit{we}\mid c)=0.0005$; when a SimPO model (Llama-3-8B-Instruct) is used, we have $p^{\text{SimPO}}(\textit{that}\mid c)=0.4326$, $p^{\text{SimPO}}(\textit{we}\mid c)=0.0008$. The log-odds $\frac{p(\textit{that}\mid c)}{p(\textit{I}\mid c)}$ is greater for UIDPO, $6.8125>6.2500$. 
Since this is a case for necessary \emph{that}, it shows UIDPO indeed has a higher chance to produce $p(\textit{that})$ than the alternative. 
However, situation is more complicated for another example where \emph{that} is redundant, $S_2:\,\textit{My boss thinks (that) I am absolutely crazy}$. UIDPO indeed assigns higher probability to the alternative than SimPO, $p^{\text{UIDPO}}(\textit{I}\mid c)=0.6303 > p^{\text{SimPO}}(\textit{I}\mid c)=0.6152$, but the redundant \textit{that} is also assigned higher probability, $p^{\text{UIDPO}}(\textit{that}\mid c)=0.0908 > p^{\text{SimPO}}(\textit{that}\mid c)=0.0782$. The log-odds $\frac{p(\textit{that}\mid c)}{p(\textit{I}\mid c)}$ is not in favor of UIDPO, $-1.9375>-2.0625$, because it indicates UIDPO still has relative higher odds to produce \emph{that} than SimPO (though the difference is small). 
With these two cherry-picked examples, we would like to intuitively demonstrate the potential effect and limit of UIDPO to produce more ``human-like'' \emph{that} complementizer. 

Finally, we use Cochran's Q + McNemar test (Repeated Measures Chi-square Test) to conduct a group-wise comparison (SFT vs. SIMPO, SFT vs. UIDPO, and SIMPO vs. UIDPO) for \Cref{tab:conjunctionthat}, and the resulting $p$-values are all below $0.001$, indicating a statitically significant effect. 
Therefore, we conclude that UIDPO-finetuned model has a higher chance to produce a necessary \emph{that} and to avoid producing a redundant \emph{that}, which aligns better with the human preference in producing uniform information density. 
Overall speaking, the effect is more salient in the former than in the latter, suggesting that it is easier to instruct a model to emphasize \emph{what should already there} not to de-emphasize \emph{what could be absent}. 
Detailed procedures and some additional analyses are provided in Appendix \ref{appendix:linguistic}.

\section{Related Work}

\paragraph{UID Theory} The Uniform Information Density hypothesis originates from the intersection of information theory and psycholinguistics, building on Shannon’s formalization of information as surprisal \citep{6773024, hale-2001-probabilistic, LEVY20081126}.
\cite{NIPS2006_c6a01432} formulates the UID hypothesis, which assumes that speakers prefer utterances that distribute information uniformly across a signal
. \cite{FLORIANJAEGER201023, speakingrationally} conducts tests to systematically discusses the mechanism of redundancy and reduction, and how they are managed by a rational speaker to uniform the information density. 
\cite{doi:10.1073/pnas.1012551108} examines that more predictable words are optimized to have shorter length and more uniform information.

Recent work has revisited and refined UID from both empirical and theoretical perspectives. \cite{meister2021revisitinguniforminformationdensity} systematically 
show that linguistic acceptability and processing difficulty correlate with deviations from uniform surprisal.
This paper opens up evaluation and application of UID in the context of large language models. Follow-up research shows that UID-inspired metrics can characterize the quality of reasoning processes \citep{gwak2026revisitinguniforminformationdensity}. 
It reports a counter-intuitive pattern of LLM reasoning trajectories: high-quality reasoning exhibits local uniformity and global non-uniformity. This observation directly inspires the redefinition of uniformity in UIDPO.

\paragraph{Preference Optimization} Reinforcement Learning from Human Feedback (RLHF) is the standard paradigm for aligning large language models with human preferences.
\citep{christiano2023deepreinforcementlearninghuman, ouyang2022traininglanguagemodelsfollow}. Despite its effectiveness, RLHF is computationally expensive and vulnerable to instability and reward mis-specification.
Recent work proposes direct preference optimization methods that remove explicit reinforcement learning. Direct Preference Optimization (DPO) \citep{rafailov2024directpreferenceoptimizationlanguage} reformulates the RLHF objective into a supervised objective by expressing the reward as a function of the log-probability difference between the policy and a reference model. This avoids training a separate reward model and stabilizes optimization
.
Subsequent work aims to further simplify and improve this framework. Simple Preference Optimization (SimPO) \citep{meng2024simposimplepreferenceoptimization} removes the reference model entirely and defines the implicit reward from the model likelihood, making the objective more consistent with the generation.
Building on this idea, later methods explore different variants of preference optimization objectives \citep{wu2025alphadpoadaptiverewardmargin, yin2025constrainalignmentsparseautoencoders}. 

\section{Conclusion}

In this work, we draw inspirations from a well-established psycholinguistic theory, Uniform Information Density, to develop a novel preference optimization algorithm, UIDPO. We compare UIDPO with SimPO, one of the State-of-the-Art preference optimization methods, and achieve comparable and sometimes better performance on major instruction following benchmarks. The key of our algorithm is to formulate the original hypothesis of UID theory as the compound of global variance and a second-order difference-based local metrics of information uniformity, which better fit the objective of preference optimization. Extensive experiments prove the necessity and superior performance of our UIDPO formulations over naïve choices. Additional linguistic analysis on complement clauses further reveals the advantages of UIDPO in obtaining language models that can distribute information in more human-aligned manners. Our work is a successful attempt to incorporate the underlying mechanisms of information processing into the post-training of LLMs that potentially better align with human language use.
 

\section*{Acknowledgments}
We sincerely thank Linyi Yang for providing insightful suggestions on model training. We thank all reviewers for giving constructive feedbacks that make this paper much improved. 
This study is funded by Shenzhen Science and Technology Program (No. JCYJ20240813094612017) and Guangdong Province ZJRC Program (No. 2024QN11X145).



\bibliography{colm2026_conference.bib}
\bibliographystyle{colm2026_conference}

\appendix

\section{More AlpacaEval 2 Results} \label{appendix:moreresults}

\begin{table}[h]
    \centering
    \begin{tabular}{lccc}
        \toprule
        \textbf{Method} & \textbf{LC(\%)} & \textbf{WR(\%)} & \textbf{Length} \\
        \midrule
        RRHF \citep{NEURIPS2023_23e6f78b} & 37.9 & 31.6 & 1700\\
        SLiC-HF \citep{zhao2023slichfsequencelikelihoodcalibration} & 33.9 & 32.5 & 1938\\
        DPO \citep{rafailov2024directpreferenceoptimizationlanguage} & 48.2 & 47.5 & 2000\\
        IPO \citep{azar2023generaltheoreticalparadigmunderstand} & 46.8 & 42.4 & 1830\\
        CPO \citep{xu2024contrastivepreferenceoptimizationpushing} & 34.1 & 36.4 & 2086 \\
        KTO \citep{ethayarajh2024ktomodelalignmentprospect} & 34.1 & 32.1 & 1878 \\
        ORPO \citep{hong-etal-2024-orpo} & 38.1 & 33.8 & 1803 \\
        R-DPO \citep{park-etal-2024-disentangling} & 48.0 & 45.8 & 1933 \\
        \midrule
        UIDPO & 49.0 & 44.8 & 1825 \\
        SimPO reproduced & 51.9 & 47.4 & 1861 \\
        \bottomrule
    \end{tabular}
    \caption{AlpacaEval 2 results of different post-training methods on Llama-3-8B-Instruct, utilizing SimPO v0.2 preference labels. All the scores are reported in SimPO paper. LC is the length-controlled win rate, and WR is the raw win rate. Length is the average token length of the responses.}
    \label{tab:morealpacaeval2}
\end{table}

Table \ref{tab:morealpacaeval2} lists more AlpacaEval 2 results retrieved from SimPO paper. The reproduced SimPO and the UIDPO are under the same training and testing setup with all other methods. That is, using princeton-nlp/llama3-ultrafeedback-armorm as the preference data and GPT-4 1106 Preview as the judge for AlpacaEval 2.

\section{MMLU results}

\begin{table}[h]
    \centering
    \begin{tabular}{ccccc}
        \toprule
        \textbf{answer extraction} & \textbf{Llama-3-8B-Instruct} & \textbf{SimPO run 1} & \textbf{SimPO run 2} & \textbf{UIDPO} \\
        \midrule
        Level 1 & \multirow{2}{*}{0.4098} & 0.3935 & 0.3910 & \textbf{0.4122} \\
        Level 2 & & 0.3983 & 0.3965 & \textbf{0.4122} \\
        \bottomrule
    \end{tabular}
    \caption{MMLU results. Level 1 \& 2 are different answer extraction mechanisms. The score of Llama-3-8B-Instruct is retrieved from the Leader-board, where only the best score is reported.}
    \label{tab:mmlu}
\end{table}

We test UIDPO and SimPO on MMLU-Pro \citep{wang2024mmlu}, and compare their performances with that of Llama-3-8B-Instruct without any post-training on the Leader-board. Since SimPO has not report its performance on MMLU-Pro, we can only compare with our reproduced SimPO. Thus we decide to include this result in the appendix instead of the main result table, to avoid confusion. According to Table \ref{tab:mmlu}, UIDPO slightly outperforms Llama-3-8B-Instruct and SimPO under both answer extraction mechanisms. Since our reproduced SimPO under-performs Llama-3-8B-Instruct, we run the benchmark for the second time, to confirm the result scores. This result shows that UIDPO can provide slight improvement on language understanding.

\section{Benchmark Statistics} \label{appendix:statistics}

\begin{table}[h]
    \centering
    \begin{tabular}{lcccccc}
        \toprule
        \multirow{2}{*}{\textbf{Method}} & \multicolumn{2}{c}{\textbf{AlpacaEval 2}} & \multicolumn{2}{c}{\textbf{AlpacaEval 2*}} & \textbf{Arena-Hard v0.1} \\
        \cmidrule(lr){2-3}\cmidrule(lr){4-5}
        & SE & avg. length & SE & avg. length & CI(\%) \\
        \midrule
        SimPO reprod. & 1.54 & 1861 & 1.75 & 1861 & -2.0/+1.8 \\
        UIDPO & \underline{\textbf{1.50}} & \underline{\textbf{1825}} & 1.75 & \underline{\textbf{1799}} & -1.9/+1.7 \\
        \bottomrule
    \end{tabular}
    \caption{Key statistics reported by main experiment benchmarks. SE is the standard error (normalized by $N-1$) of the win rate. Avg. length is the average token length of responses. CI is the non-overlapping confidence interval with the baseline.}
    \label{tab:resultstats}
\end{table}

UIDPO has slightly better statistics than SimPO. Table \ref{tab:resultstats} lists some key statistics reported by each benchmark. A smaller standard error implies more stable performance.
Shorter average length means that UIDPO is less likely to exploit length bias. It is also a preferable feature in application, in terms of both user experience and cost. The Confidence Interval (CI) reported by Arena-Hard measures how similar a model's responses are to the baseline (GPT-4 0314 in this case). It is not directly related to performance.

\section{Average Log Probability of Chosen Responses in $\gamma$ Ablations} \label{appendix:logpgamma}

\begin{figure}
    \centering
    \includegraphics[width=.8\linewidth]{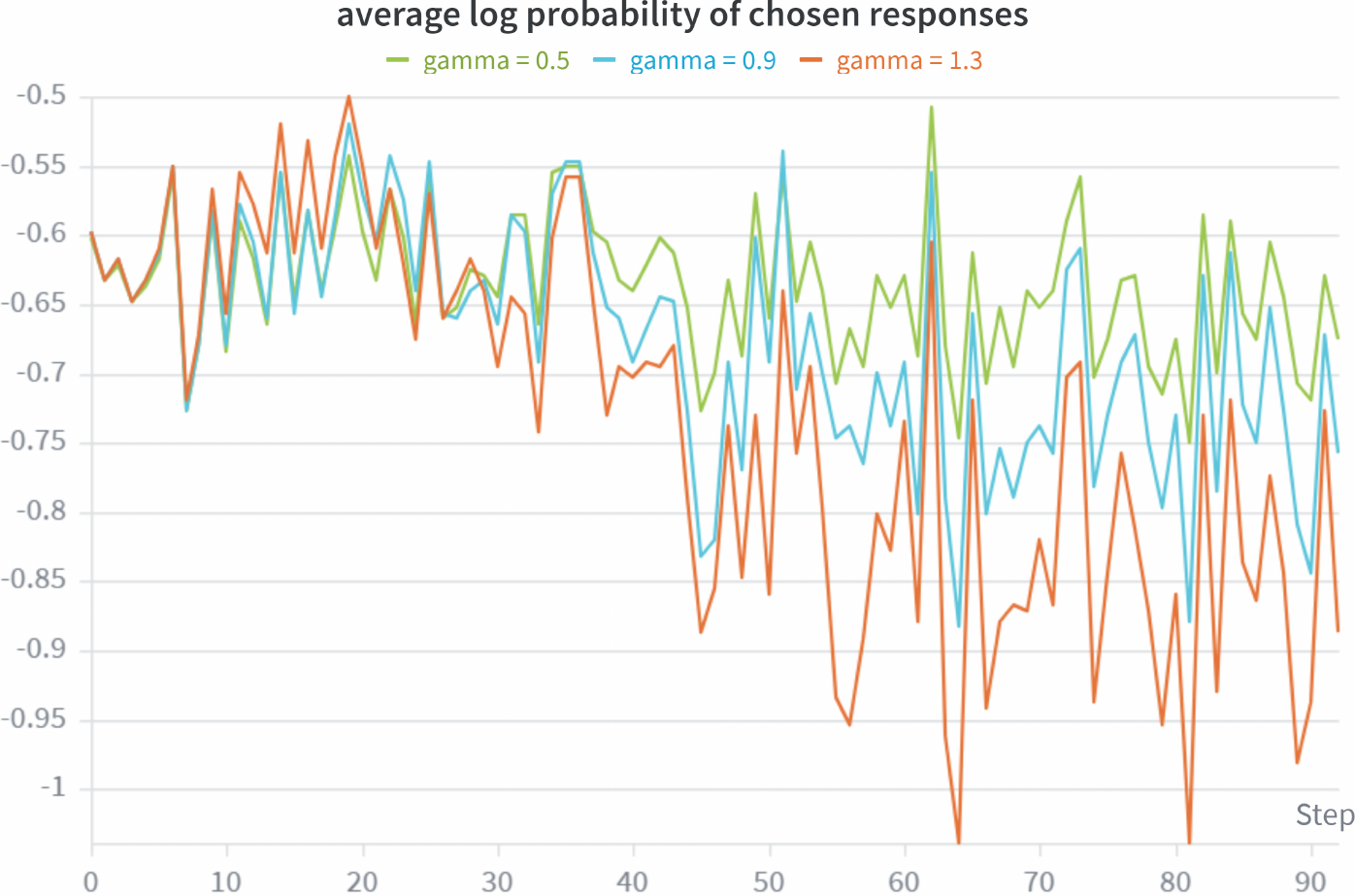}
    \caption{Average log probability of chosen responses during training, when $\gamma = 0.5, 0.9, 1.3$.}
    \label{fig:logpgamma}
\end{figure}

Figure \ref{fig:logpgamma} displays three examples of the average log probability curves during training. We select three curves of evenly spaced $\gamma$s for clearer exhibition, and other curves also conform the same pattern. As $\gamma$ becomes larger, the average log probability drops more sharply. This confirms the discovery in SimPO paper, that a larger $\gamma$ flattens the probability distribution.

\section{Reward Margins and UID Margins}

\begin{figure}
    \centering
    \includegraphics[width=0.8\linewidth]{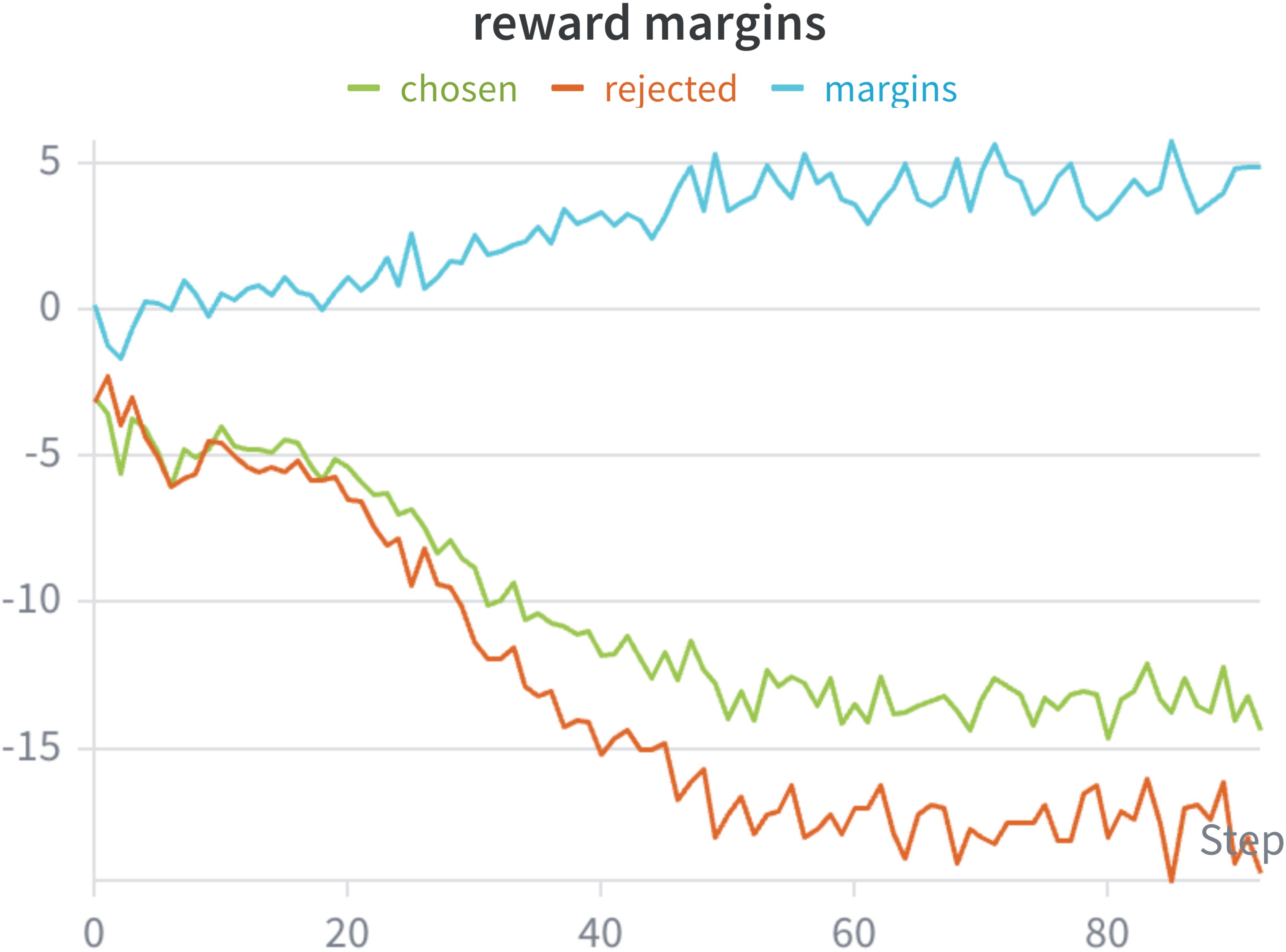}
    \caption{The reward margins between chosen responses and rejected responses during training, under the best performing setup.}
    \label{fig:rewardmargin}
\end{figure}

\begin{figure}
    \centering
    \includegraphics[width=0.8\linewidth]{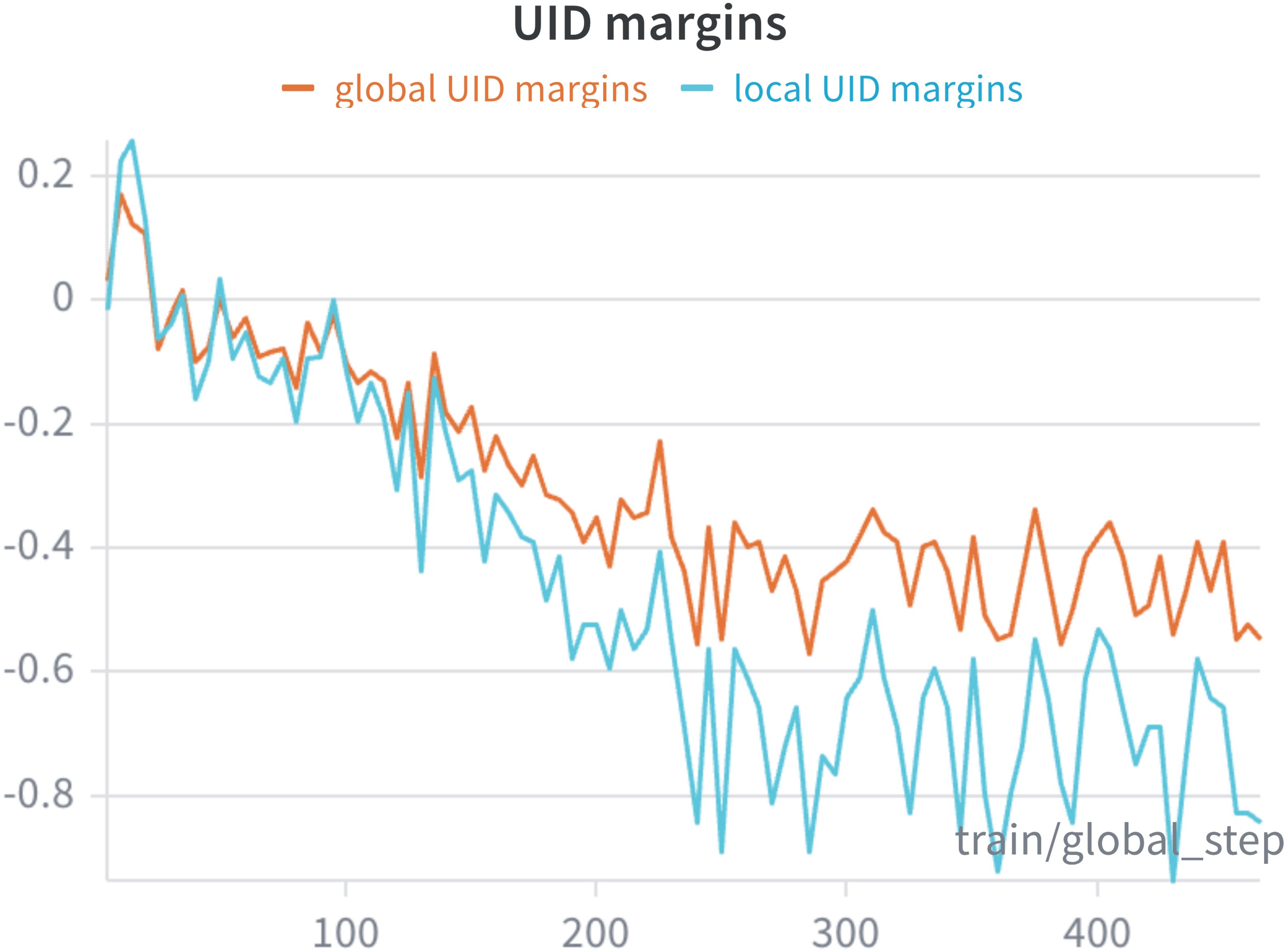}
    \caption{The global and local UID margins between chosen responses and rejected responses during training, under the best performing setup.}
    \label{fig:uidmargin}
\end{figure}

To exhibit the approximate numerical range and the trend of the UIDPO reward and UID terms, we plot the following two figures. Figure \ref{fig:rewardmargin} displays the widening gap between rewards of chosen responses and rejected responses. Figure \ref{fig:uidmargin} displays the dropping UID, where the global term drops but slower. This is probably due to the choice of $\alpha$. More weight on the local term (negative contribution) forces the global term to drop, while the positive contribution of the global term resists. this causes the widening gap between the global UID and the local UID.

\section{Linguistic Analysis Details} \label{appendix:linguistic}

This section provides additional details about the complement-clause
analysis in \Cref{sec:Experiments}. All three models in this analysis
use the Llama-3-8B-Instruct backbone.

\paragraph{Corpus and clause extraction.}
We use the fluent portion of the parsed Switchboard corpus
\citep{godfrey1992switchboard}. Candidate clauses are extracted using
an NLTK-based approximation of the \texttt{tgrep2} query used by
\citet{FLORIANJAEGER201023}. The procedure selects trees containing an
\texttt{SBAR} associated with a verbal projection and requires the
clause to begin with either an overt complementizer \textit{that} or
a Treebank null-complementizer annotation. It also removes clauses
with a subject gap, extraposed clauses, clauses containing a
wh-expression, and common parenthetical uses such as \textit{I mean}
and \textit{you know}.

The structural constraints are adapted from the following
\texttt{tgrep2} query:

\begin{verbatim}
/^sbar/ $,, /^vb/
[<(in<"that") |< "-none-"]
<(/^s/ < (/-sbj/ !< "-none-"))
!<(/^n/ < ! /^'0,1[a-zA-Z]+.*/)
!</^wh/
\end{verbatim}

Our NLTK implementation approximates these constraints with the
following procedure:

\begin{verbatim}
for each parsed tree:
    require at least one SBAR and one VP
    require overt "that", null "0", or a -NONE- node
    reject trees containing an empty subject
    reject extraposed clauses identified by an indexed NP
    reject WH constituents and clauses headed by "whether"
    reject short parenthetical "I mean" or "you know" uses
    retain the remaining tree
\end{verbatim}

The filtering procedure retains 4,867 of the 177,804 parsed trees.
Within these trees, we identify 4,817 complement-clause instances:
962 with an overt \textit{that} and 3,855 with a null complementizer.
These instances correspond to 1,768 distinct prefixes. The descriptive
statistics below treat clause occurrences, rather than unique
prefixes, as separate observations.

\paragraph{Construction of model inputs.}
For an overt complementizer, we retain the observed \textit{that}.
For a null complementizer, we insert \textit{that} at the annotated
clause boundary. We then retain the prefix preceding this position and
the first token $w$ of the embedded clause. This construction allows
\textit{that} and $w$ to be evaluated as alternative continuations of
the same prefix $c$.

\paragraph{Probability-based measures.}
For each model, we compute the negative log likelihoods
$
    \operatorname{NLL}_{\theta}(\textit{that}\mid c)
    = -\log p_{\theta}(\textit{that}\mid c)
$
and
$
    \operatorname{NLL}_{\theta}(w\mid c)
    = -\log p_{\theta}(w\mid c)
$.
The latter is the probability of beginning the embedded clause
without an overt complementizer; it is \textbf{not} the probability of $w$
conditioned on \textit{that}. We also count the number of vocabulary
items assigned a higher probability than \textit{that} under the same
prefix. A value of zero therefore indicates that \textit{that} is the
model's most probable continuation. We report the mean of this
rank-based measure below.

\paragraph{Descriptive results.}
UIDPO assigns a higher NLL to \textit{that} than SFT on 85.3\% of the
analyzed instances. Across all instances, the mean ratio between the
UIDPO and SFT NLLs is 1.070. Conversely, UIDPO assigns a lower NLL than
SimPO on 96.5\% of the instances, with a mean UIDPO-to-SimPO NLL ratio
of 0.945.

The mean difference

\begin{equation}
    \operatorname{NLL}_{\theta}(\textit{that}\mid c)
    - \operatorname{NLL}_{\theta}(w\mid c)
\end{equation}

is $-2.39$ for SFT, $-2.27$ for UIDPO, and $-2.13$ for SimPO.
Thus, all three models assign a higher probability, on average, to
\textit{that} than to beginning the embedded clause directly.

For the rank-based measure defined above, the average numbers of
vocabulary items assigned a higher probability than \textit{that} are
46.18 for SFT, 59.91 for UIDPO, and 63.28 for SimPO. As a qualitative
complement, we also count the tokens that most frequently rank above
\textit{that} across the analyzed contexts. Although their exact order
varies slightly across models, the same ten items occur most
frequently for SFT, SimPO, and UIDPO: \textit{I}, \textit{it},
\textit{you}, \textit{the}, comma, period, \textit{to}, \textit{we},
\textit{in}, and \textit{there}.

Taken together, these aggregate statistics place UIDPO between SFT
and SimPO in its overall propensity to produce \textit{that}. They are
descriptive and, on their own, do not establish whether a model
distinguishes contexts that favor realization from those that favor
omission.

\end{document}